\documentclass[letterpaper, 10 pt, conference]{ieeeconf}

\IEEEoverridecommandlockouts            % This command is only needed if 
\usepackage{amssymb}  % assumes amsmath package installed
\usepackage{graphicx} % for pdf, bitmapped graphics files
\usepackage{xcolor}
\usepackage{multirow}
\usepackage{multicol}
\usepackage{booktabs}
\usepackage[utf8]{inputenc}
\usepackage{amsmath,amsfonts,booktabs,cite} % general useful stuff
\usepackage{siunitx,textcomp} % for the degree symbol
\usepackage[hidelinks]{hyperref} % to have hyperlinks and for the \ref macros
\usepackage{subcaption}
\let\labelindent\relax
\usepackage[inline]{enumitem} % inline enumerate
\usepackage[nolist]{acronym}  % acronyms by the \ac{label} command

\newacro{slam}[SLAM]{Simultaneous Localization and Mapping}
\newacro{icp}[ICP]{Iterative Closest Point}
\newacro{tsdf}[TSDF]{Truncated Signed Distance Function}
\newacro{esdf}[ESDF]{Euclidean Signed Distance Field}
\newacro{pcn}[PCN]{Point Completion Network}
\newcommand{\figref}[1]{\hyperref[#1]{Fig.~\ref*{#1}}}
\newcommand{\tabref}[1]{\hyperref[#1]{Tab.~\ref*{#1}}}
\newcommand{\secref}[1]{\hyperref[#1]{Sec.~\ref*{#1}}}
\newcommand{\algoref}[1]{\hyperref[#1]{Alg.~\ref*{#1}}}

\newcommand{\subfigref}[1]{(\subref{#1})}

\newcommand{\matr}[1]{\mathbf{#1}}

\DeclareMathOperator*{\argmin}{arg\,min}

\def\hyper{\mathcal{M}}
\def\map{\matr{M}}
\def\voxb{\matr{G}}
\def\sem{\matr{S}}
\def\repl{\matr{O}}
\def\augm{\matr{A}}

\def\fscore{F\textsubscript{1}-score}
\def\turtle{TurtleBot 3 Waffle}
\def\bestcolor{(best viewed in color)}
\def\sota{state-of-the-art}
\def\ie{\textit{i.e.},}
\def\eg{\textit{e.g.},}
\def\etal{\textit{et al.}}
\def\pc{point-cloud}

\title{\LARGE \bf 
    Augmented Environment Representations with Complete Object Models
}

\author{Krishnananda~Prabhu~Sivananda, Francesco~Verdoja, Ville~Kyrki%
\thanks{K.~P.~Sivananda is with Cargotec Oy, F.~Verdoja and V.~Kyrki are with
the School of Electrical Engineering, Aalto University, Finland.}
\thanks{The work was conducted while K.~P.~Sivananda was a student at the School
of Electrical Engineering, Aalto University, Finland.}
\thanks{For contacts: \texttt{francesco.verdoja{@}aalto.fi}}}

\begin{document}

\maketitle

\thispagestyle{empty}
\pagestyle{empty}

%%%%%%%%%%%%%%%%%%%%%%%%%%%%%%%%%%%%%%%%%%%%%%%%%%%%%%%%%%%%%%%%%%%%%%%%%%%%%%%%

\begin{abstract}
While 2D occupancy maps commonly used in mobile robotics enable safe navigation
in indoor environments, in order for robots to understand and interact with
their environment and its inhabitants representing 3D geometry and semantic
environment information is required. Semantic information is crucial in
effective interpretation of the meanings humans attribute to different parts of
a space, while 3D geometry is important for safety and high-level understanding.
We propose a pipeline that can generate a multi-layer representation of indoor
environments for robotic applications. The proposed representation includes 3D
metric-semantic layers, a 2D occupancy layer, and an object instance layer where
known objects are replaced with an approximate model obtained through a novel
model-matching approach. The metric-semantic layer and the object instance layer
are combined to form an augmented representation of the environment. Experiments
show that the proposed shape matching method outperforms a \sota{} deep learning
method when tasked to complete unseen parts of objects in the scene. The
pipeline performance translates well from simulation to real world as shown by
\fscore{} analysis, with semantic segmentation accuracy using Mask R-CNN acting
as the major bottleneck. Finally, we also demonstrate on a real robotic platform
how the multi-layer map can be used to improve navigation safety.
\end{abstract}

\section{Introduction}
\label{sec:intro}

In recent years, we have witnessed great advances in computer vision which have
given robots the ability to understand the world around them like never before
\cite{heMaskRCNN2017, yuanPCNPointCompletion2018}. At the same time, advances in
control of mobile platforms have enabled robots from quadrupedal robot dogs to
drones to move in different environments and react to unpredicted circumstances
with high reliability. 

\begin{figure}
    \centering
    \includegraphics[width=\linewidth]{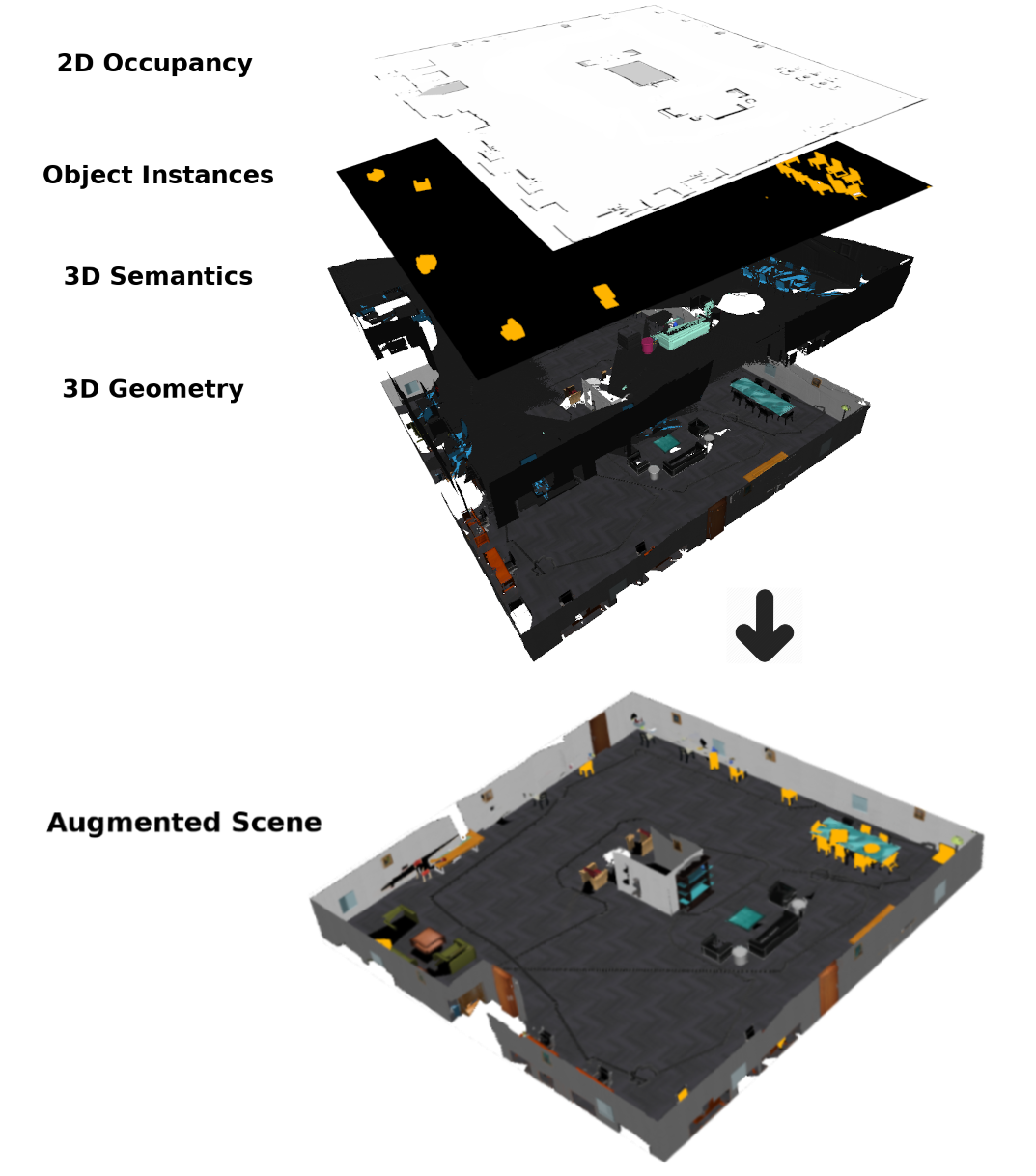}
    \caption{An environment representation built using the proposed
    framework.}\label{fig:map}
\end{figure}

However, when considering a mobile robot’s ability to perform advanced
high-level tasks in unstructured human-inhabited environments, there is a
crucial bottleneck stifling the ability of these sensing and acting
breakthroughs to lead to the desired level of autonomy, performance, and
interaction: the way robots reason on this rich perceptual knowledge is still
very limited. When considering navigation for example, ground robots still rely
on 2D occupancy maps built using \ac{slam} algorithms for path planning and
obstacle avoidance. These maps are built using only laser scanners and the rich
information coming from the variety of visual sensors available to modern robots
is mostly ignored. The combination of these sensors would allow the robot to
formulate a better understanding of its surroundings and, in the case of
navigation, to take more intelligent choices to plan its route, \eg{} by
avoiding obstacles invisible to the laser scanner. Similarly, while interacting
with humans, understanding object semantics is crucial for understanding
commands in everyday language \cite{pandey2014towards}; having a richer
understanding of the environment opens the door for complex capabilities such as
mobile manipulation and natural user interaction.

Though methods exist in literature to construct a meaningful 3D representation
of the environment \cite{newcombeKinectFusionRealtimeDense2011,
oleynikovaVoxbloxIncremental3D2017, mccormacSemanticFusionDense3D2017}, the
acquisition of a full 3D map of an environment is a time-consuming process. In
most realistic scenarios, the underlying geometry reconstructed by these methods
is often incomplete, due to time concerns, occlusions, or environmental
traversability limitations affecting the perception of the scene during the
mapping process. However, partial geometry is a considerable drawback in some
applications; for example, when a robot has to interact with an object, the lack
of knowledge of complete geometry can pose safety risks either to the object,
the surroundings, or the robot itself. Moreover, complete objects model are also
essential for user immersion and safe navigation in telepresence applications
\cite{cesta2012into}. 

In this paper, we propose a mapping pipeline to construct 3D metric-semantic
environment representations which include estimates of the full extent of
objects in the environment through object shape completion. An example of this
representation is shown in \figref{fig:map}. We also propose a novel
deterministic approach based on matching (and potential replacement) of a
partial object with a similar synthetic model from a known database, which
avoids the unpredictability of recent deep-learning-based shape completion
approaches. In the experiments, we validate our approach by comparing it to a
\sota{} deep-learning-based approach for shape completion of chairs. 

To summarize, the contributions of this paper are:

\begin{itemize}
    \item A pipeline  to generate a 3D metric-semantic mesh representation of an
    environment that completes partial objects with similar complete synthetic
    models.
    \item Design and evaluation of a novel method to compare and match partial
    observations of 3D object instances against a database of synthetic models
    and to replace the partial 3D mesh with the matched model in the
    reconstructed 3D mesh of the environment.
    \item Evaluation and comparison of a learning-based approach to complete
    partial observations of 3D object instances against our proposed method.
    \item A demonstration of benefits of the proposed environment representation
    in a 2D navigation case on a real \turtle{}.
\end{itemize}

\section{Related Work}
\label{sec:related}

Building environment representations where complete models of objects are
estimated requires first reconstructing the 3D geometry of the environment,
understanding its semantics, and then shape-completing each object instance. In
this section, we will review \sota{} relevant to these fields, together with
relevant works in multi-layer mapping for robots.

\paragraph*{Reconstruction of 3D geometry} 
The availability of inexpensive RGB-D cameras such as Kinect has resulted in
profound advances in developing 3D reconstruction methods. KinectFusion
\cite{newcombeKinectFusionRealtimeDense2011} popularised \ac{tsdf}-based
reconstructions as capable of real-time reconstruction of small environments.
The main drawback of KinectFusion is the use of a fixed-size voxel grid which
requires the map size to be known and a large amount of memory. Multiple
extensions have been proposed to address these shortcomings \cite{Whelan12rssw,
article}. Among these, Nießner \etal{} \cite{10.1145/2508363.2508374} propose a
spatial hashing scheme to compress space while allowing real-time access and
updates of underlying surface data. This spatial hashing was faster than the
hierarchical grid data structure used in other methods, however, it relies
heavily on GPUs for real-time performance. By exploiting the spatial hashing
technique employed in \cite{10.1145/2508363.2508374}, Voxblox
\cite{oleynikovaVoxbloxIncremental3D2017} focused on improving memory efficiency
and real-time performance on the CPU by building incremental \acp{esdf} from
\acp{tsdf}. This makes Voxblox able to reconstruct large scale scenes with
reasonable computational costs. In this work, since we are interested in
building 3D model of complete environments, we employ Voxblox in the geometric
reconstruction module.

\paragraph*{Semantic reconstruction}
The recent innovations in image segmentation methods enabled semantic knowledge
to be incorporated into 3D reconstructions into so called semantic-metric
representations. Among the methods to perform 3D semantic mapping,
SemanticFusion \cite{mccormacSemanticFusionDense3D2017} combines the real-time
SLAM system ElasticFusion \cite{Whelan2015ElasticFusionDS} and a CNN for object
detection with a Bayesian update scheme for semantic label integration,
Voxblox++ \cite{grinvaldVolumetricInstanceAwareSemantic2019}, an extension of
Voxblox, combines geometric segmentation of depth data with instance
segmentation by Mask R-CNN \cite{heMaskRCNN2017}. A data association strategy
keeps track of labels to ensure global consistency. On the same line,
Kimera-Semantics \cite{rosinolKimeraOpenSourceLibrary2020} uses Voxblox allowing
however flexibility for choice for the 2D instance segmentation algorithm. It
runs faster than \cite{grinvaldVolumetricInstanceAwareSemantic2019} and shows
accurate reconstruction when compared against ground-truth. In this work, we
employ Kimera-Semantics with Mask R-CNN to build semantic representations.

\paragraph*{Shape completion}
Shape completion is the process for inferring a full 3D model of an object based
on a partial measurement. Two main approaches to the problem exist:
model-matching and generative. Model-matching methods attempt at matching a
partial view to a similar model. The most recent example of this class of
methods is Scan2CAD \cite{Avetisyan_2019_CVPR} which uses a 3D CNN trained on a
custom dataset to learn joint embedding between real and synthetic 3D models to
predict accurate correspondence heatmaps between model and scan. It requires
entire 3D scan of a scene as input and tries to find the alignment of models of
all the objects in the scene. The method, however, does not utilize any semantic
information of the scene. Recently, generative deep learning methods for shape
completion have attracted increasing interest in the community. Although methods
working on voxel data exist \cite{daiShapeCompletionUsing2017}, \pc{}s are
better suited to represent objects at different resolution. Among the methods
working on \pc{}s \cite{huangPFNetPointFractal2020a, DBLP:conf/iclr/ChenCM20,
Tchapmi_2019_CVPR}, one of the best performing ones is \ac{pcn}
\cite{yuanPCNPointCompletion2018}, a shape completion method based on the
PointNet architecture. It is capable of generating high-resolution completions
and shows generalization over unseen objects and real-world data. However, most
of the results presented in works on deep learning shape completion only pertain
synthetic data and their ability to transfer from synthetic-partial data to real
data has been questioned \cite{DBLP:conf/iclr/ChenCM20}. Moreover these methods
often do not account for much input noise while training. The problem of scene
completion (\ie{} shape-completing all objects in a scene) is less explored and
in that context, which we target in this work, errors in semantic segmentation
or geometric reconstruction of the scene can further increase the transfer gap
and lessen the performance of deep learning methods even more. In this work, we
explore the ability of \ac{pcn} to transfer to noisy real data as well as
propose a novel model-matching approach for scene completion and augmentation.

\begin{figure*}
    \centering
    \includegraphics[width=\linewidth]{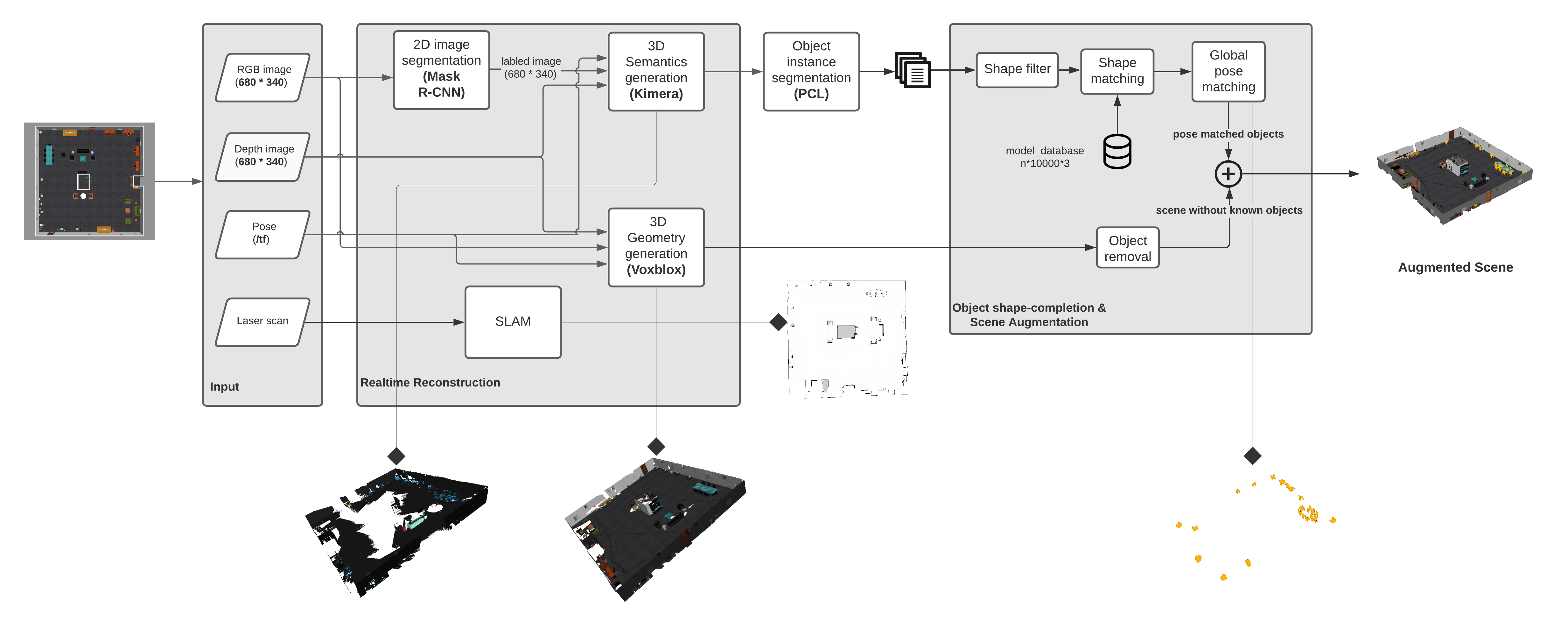}
    \caption{Detailed overview of the pipeline}\label{fig:full_pipe}
\end{figure*}

\paragraph*{Multi-layer mapping}
Multi-layer maps for robotics are not a new concept. Many works have proposed
hierarchical mapping methods where a 2D occupancy map is maintained together
with traversability graphs and topology \cite{zender_conceptual_2008,
crespoRelationalModelRobotic2017}. However, these architectures rarely include
multiple maps capturing the environment as seen by different sensor modalities
but rather abstract the environment to facilitate reasoning and human-robot
interaction.

Recently, motivated by the advancements in computer vision and machine learning,
interest has increased toward multi-layer mapping formalisms able to include
different sensor information. In \cite{zaenkerHypermapMappingFramework2019} a 2D
multi-layer mapping framework composed of a metric, semantic, and exploration
layers is proposed, and its application in the context of autonomous semantic
exploration is presented. In that work, the focus is left on 2D mapping only,
and no object shape completion is attempted. Rosinol \etal{}
\cite{rosinol3DDynamicScene2020} propose a hierarchical graph comprised of
multiple layers, including one where object extents are estimated by fitting a
CAD model to the partial object. In this work, we propose a more general
approach to shape completion, by matching over a database of object models,
instead of a single one.

\section{Problem formulation}
\label{sec:prob}

The aim of this work is to build rich map representations where geometric
information of the environment is maintained together with the object semantics.
Additionally, we want to maintain an estimate over the full extent of each
object.

Formally, we want to build a hierarchical multi-layer map $\hyper{} = \{\map{},
\voxb{}, \sem{}, \repl{}\}$, composed of the following layers:
\begin{enumerate}
    \item 2D geometry $\map{}$: a representation of traversable area, often used
    for navigation and obstacle avoidance;
    \item 3D geometry $\voxb{}$: a representation of the environment where
    objects can be recognized by their appearance;
    \item 3D semantics $\sem{}$: a representation where semantic information
    over objects in the environment is maintained;
    \item Object instances $\repl{} = \{o^*\}$: for each object in the
    environment, we want to estimate its full extent $o^*$ from a partial view
    $\tilde{o}$.
\end{enumerate}

In the next section we will discuss the proposed pipeline, which builds all the
aforementioned maps iteratively by means of visual and range information
obtained by a robotic platform exploring the environment.
\section{Method}
\label{sec:method}

The proposed pipeline, shown in \figref{fig:full_pipe}, is split into
\emph{Realtime Reconstruction} and \emph{Scene Augmentation}. \emph{Realtime
Reconstruction} includes modules to build the geometric and semantic
representations. These reconstructions form the input to the second half of the
pipeline, that performs \emph{Scene Augmentation} on the input to deliver the
final geometric representation of the scene having objects with complete
geometry. All representations are kept aligned to a 2D occupancy map $\map{}$
built through \ac{slam}. The proposed pipeline consists of the combination of
existing software components along with novel components to produce the final
representation. While developing the pipeline, the emphasis was kept on the flow
of data rather than on the individual component implementations. Hence, each
component is independent from the others, thereby easily replaceable with any
current or future work that can ensure the same data flow. 

\subsection{Realtime reconstruction}
\label{sec:recon}

The purpose of the reconstruction module is to generate the 3D geometric
representation  $\voxb{}$ and the semantic-geometric representation $\sem{}$
online. Both layers will be represented as 3D colored meshes; in $\voxb{}$, each
face's color will represent the color information obtained from the camera,
while in $\sem{}$ the color will map to a semantic object class. Semantic
information is gathered through a deep learning based pixel-level instance
segmentation method, the output of which along with the respective depth image
and pose information is used to generate the semantically annotated 3D mesh. In
this work, we use Voxblox \cite{oleynikovaVoxbloxIncremental3D2017} and
Kimera-Semantics \cite{rosinolKimeraOpenSourceLibrary2020} to construct the
geometric and semantic-geometric representation respectively, employing Mask
R-CNN \cite{heMaskRCNN2017} as the deep learning algorithm to obtain the object
semantic segmentation from images.

This stage outputs $\voxb{}$ and $\sem{}$, \ie{} a geometric and a
semantic-geometric representation of the environment available in a mesh format
which are converted to \pc{}s due to ease of processing for the latter stage of
the pipeline.

\begin{figure}
	\centering
	\begin{subfigure}[b]{.49\linewidth}
	    \centering
		\includegraphics[width=\linewidth]{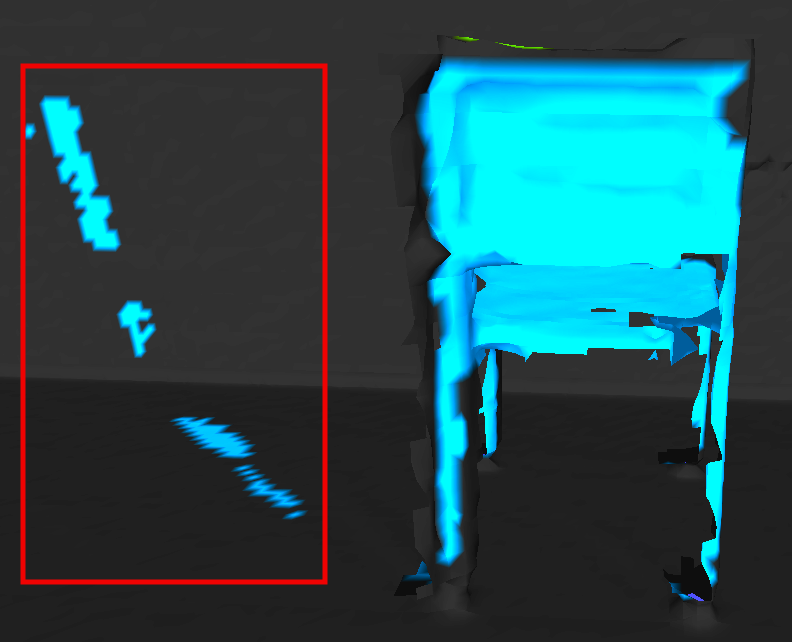}
		\caption{Spread onto wall and floor}\label{fig:wall}
	\end{subfigure}
	\begin{subfigure}[b]{.49\linewidth}
	    \centering
		\includegraphics[width=\linewidth]{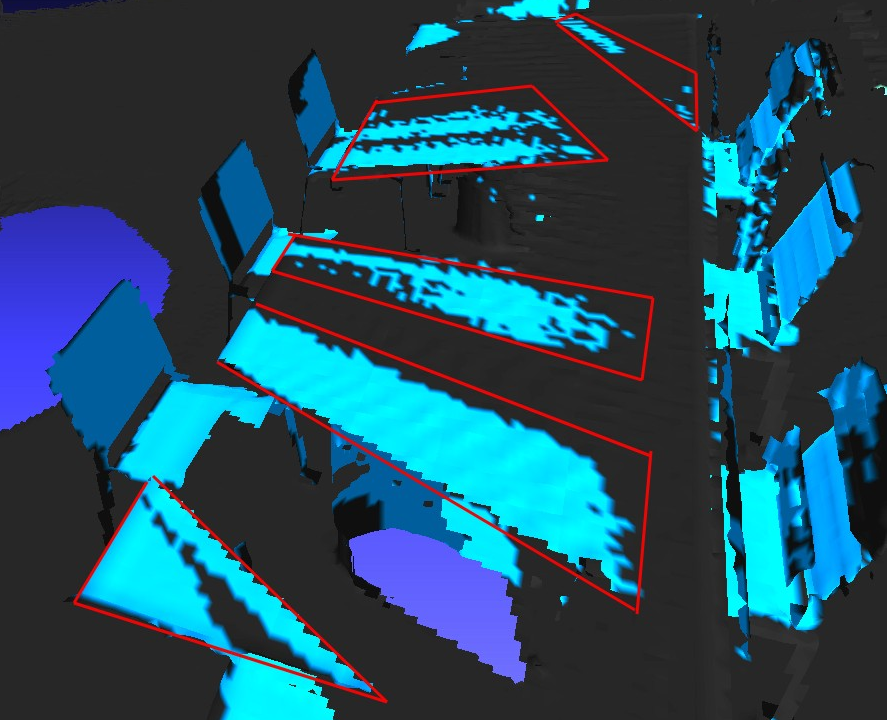}
		\caption{Spread onto a table}\label{fig:table}
	\end{subfigure}
    \caption{Discrepancies caused by inaccurate object mask. In both pictures
    the marked regions represent erroneous spread of a label
    \bestcolor.}\label{fig:maskrcnn_desc}
\end{figure}

\subsection{Object shape completion and scene augmentation}
\label{sec:completion}

In order to obtain the object instances to shape-complete, the semantic point
cloud is segmented into individual clusters, with each cluster representing a
partial view $\tilde{o}$ of an object $o$. Object instances are identified using
differences of normals. Then, given a semantic class $c$, for each partial view
$\tilde{o}$ for which $\sem{}(\tilde{o}) = c$, we propose an object
shape-completion method $f_c$ to estimate its full extent $o^*$. Formally, $f_c:
\tilde{o} \to o^*$. In this work, we demonstrate the augmentation by replacing
chairs, but the proposed methodology can be applied to other classes of objects
as well.

\subsubsection{Shape Filter}
\label{sec:filter}

After clustering, it can happen that some clusters $\tilde{o}$ may be
incorrectly classified to class $c$. This can happen \eg{} for issues of time
synchronization between RGB and depth images, or because the mask obtained from
Mask R-CNN may contain parts that do not belong to the object. The marked
regions in \figref{fig:maskrcnn_desc} are examples of such cases. To avoid
completing the shapes of mislabeled objects, we pass each cluster through two
filters aiming at recognizing erroneous labeling. 

The first filter aims at recognizing label-bleeding along walls and floor by
identifying planarity of a cluster \pc{} by its covariance matrix. A cluster is
discarded if the covariance matrix shows values close to zeros for at least one
of the axes.

The second filter aims at identifying mislabeled objects. Objects of a certain
class tend to have similar shape and size, and these can be used to identify
objects which do not conform to the class model. In practice, we determined the
distance of the farthest point to the object centroid $\lambda$ and removed
objects for which this measure was out of range for any sort of chair commonly
found in an office or home environment. Despite the filtering, some false
positives may remain for later stages. 

\subsubsection{Shape matching}

In order to estimate the full extent of an object, we match its partial shape
$\tilde{o}$ to a similar synthetic model $o^*$. Formally, given $\tilde{o}$, a
partial view of an object of class $c$, and $\matr{C} = \{o_i\}_{i=1}^N$, a
database of $N$ object models $o_i$ of class $c$, we want to find 
\begin{equation}\label{eq:match}
    o^* = \argmin_{o_i \in \matr{C}} \delta(o_i, \tilde{o})
\end{equation}
for some distance metric $\delta$. In this study, a custom \pc{} database of
chairs was created from the 6778 3D models available in the ShapeNet dataset
\cite{shapenet2015}.

In order to compute the distance $\delta$, $o_i$ and $\tilde{o}$ need to first
be registered to the same pose.

\begin{figure*}
    \centering
    \begin{subfigure}[b]{.37\linewidth}
        \centering
        \includegraphics[height=5.5cm]{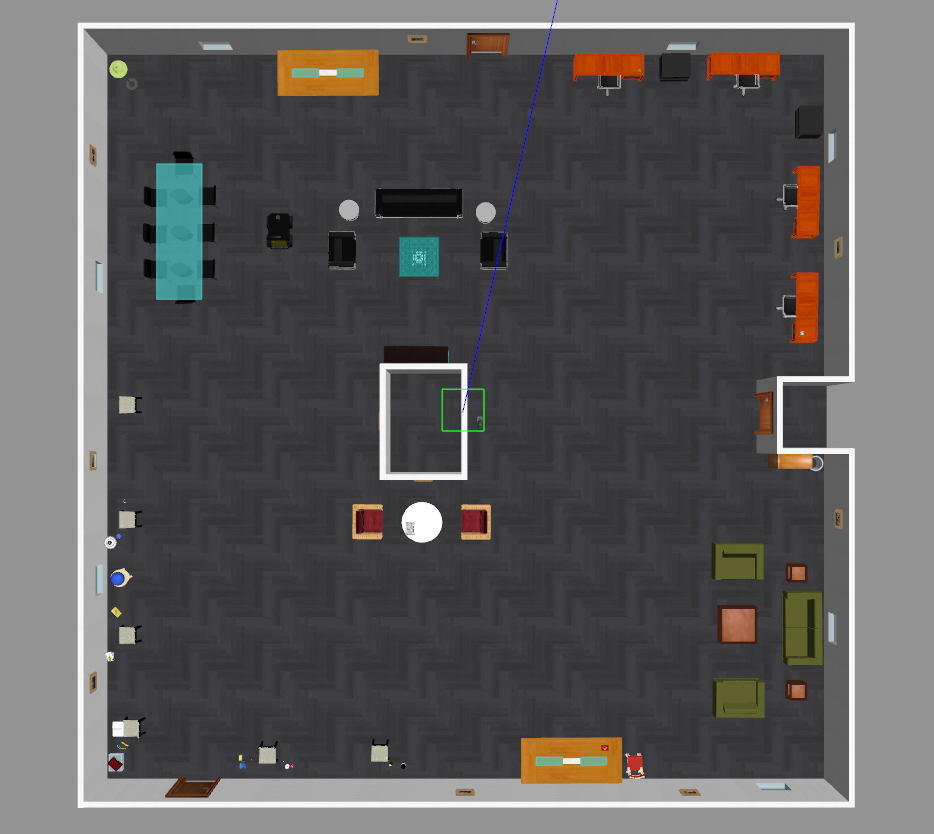}
        \caption{Top view from Gazebo}
        \label{fig:office}
    \end{subfigure}
    \begin{subfigure}[b]{.37\linewidth}
	    \centering
        \includegraphics[height=5.5cm]{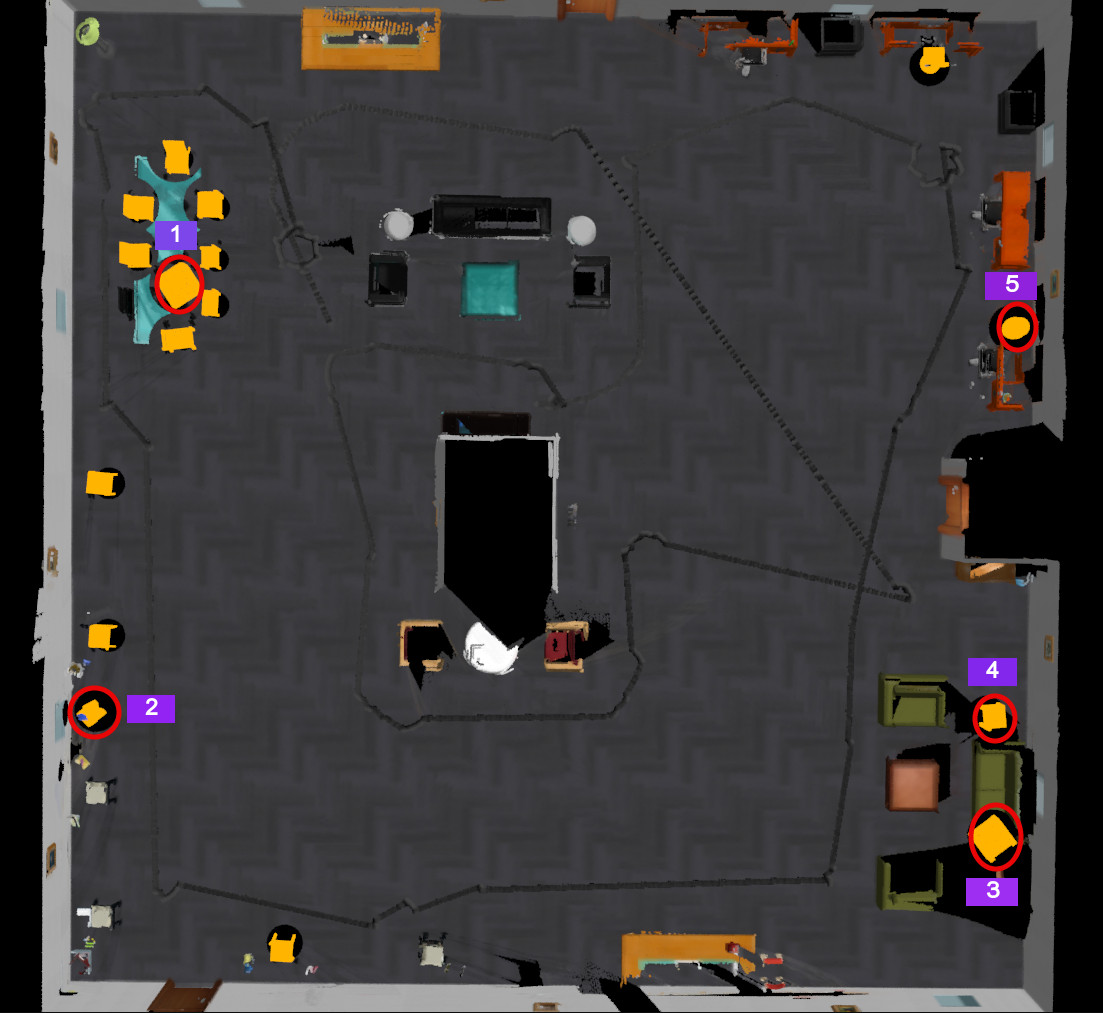}
		\caption{Augmented scene}\label{fig:false_pos}		
	\end{subfigure}
	\begin{subfigure}[b]{.24\linewidth}
	    \centering
		\includegraphics[height=5.5cm]{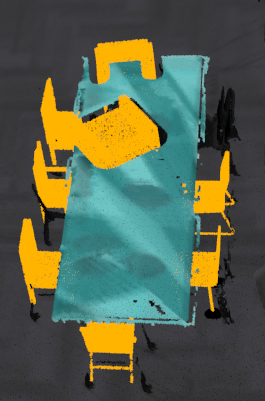}
		\caption{Meeting table detail}\label{fig:mm_tab}		
	\end{subfigure}
    \caption{Simulated office scene \subfigref{fig:office}. In the 3D geometric
    model \subfigref{fig:false_pos}, chair partial observations have been
    replaced with the complete models and each area marked in red in
    \subfigref{fig:false_pos} denotes a false positive. A detail of the meeting
    table in the North-West corner of the environment is shown zoomed in
    \subfigref{fig:mm_tab} \bestcolor{}.}\label{fig:performance}
\end{figure*}

\subsubsection*{Pose matching}
\label{sec:pose}

Inferring the pose of an object directly from its raw partial \pc{}
representation is quite difficult. Instead, it is easier to find a
transformation that can be applied to the model that will align it with the
object through \pc{} registration methods through \ac{icp}. However, \ac{icp} is
a computationally expensive algorithm and is susceptible to getting stuck at
local minima, so it is best if any initial estimate of the transformation is
available to loosely align the source (database models) with the target (partial
\pc{}s). \ac{icp} can then fine-tune the coarse transformation in very few
iterations to tighten the alignment. Given that the natural pose of any grounded
objects will always be around the $z$-axis, to find a coarse transformation, a
random model from the dataset is selected and a set of uniformly sampled
rotations are performed around its $z$-axis. The transformation matrix for the
rotation yielding the minimum average point-to-point distance is used to
initialize the \ac{icp} algorithm. \ac{icp} fine-tunes this rotation over the
$z$-axis to provide the final transformation matrix. The advantage of this
approach is that it is applicable to any grounded object.

\subsubsection*{Model matching}
\label{sec:match}

Finding a match involves searching through the database. The final
transformation refined after \ac{icp} method is applied to each model and the
corresponding point-to-point distance with the partial \pc{} is recorded. This
is used as a measure of model distance $\delta$. The hypothesis is that the
model that most closely resembles the actual object should return the smaller
distance. The matched model $o^*$ in the local frame of the partial object is
re-scaled and translated back into the world frame and added to the object
representation layer $\repl{}$.

\subsubsection{Scene augmentation}

Finally, after having built $\voxb{}$, $\sem{}$, and $\repl{}$, we build an
augmented scene, \ie{} a representation where we replace objects in $\voxb{}$
with their complete counterparts in $\repl{}$. While not constituting one of the
layers of $\hyper$, this additional \emph{virtual} representation can be used
for planning and navigation, as we will show in \secref{sec:nav}. To this end,
we construct a merged \pc{} $\augm{} = (\voxb{} \setminus \matr{S}) \cup
\repl{}$, where $\matr{S} \subset \voxb{}$ is the subset of points of $\voxb{}$
that are within a point-to-point distance threshold $\epsilon$ from any of the
object models in $\repl{}$.
\section{Experiments}
\label{sec:exp}

In this section, we present experiments aiming to validate the proposed method
by answering the following questions:
\begin{enumerate}
    \item is the pipeline able to reconstruct environments while navigating in
    them?
    \item when considering chairs, how reliably can the proposed method locate
    them and estimate their full extent?
    \item how does the object extent estimated by the proposed method compare
    with \sota{} shape-completion deep learning methods?
\end{enumerate}
In order to answer these questions we conducted multiple experiments in
simulation (\secref{sec:sim}) and on real robotic platforms (\secref{sec:real}).

As a measure to evaluate the method's ability to recognize objects, we used the
\fscore{}, 
\begin{equation}
	F_1 = 2\frac{Precision \cdot Recall}{Precision+Recall}.
\end{equation}

All experiments have been run on a laptop with Intel Core i7-8750H, 2.20GHz CPU,
Nvidia GeForce 1050 Ti GPU, and 16 GB RAM. While they are not presented here due
to space limitations, we conducted more experiments that can be found in
\cite{kpthesis2020}.

Finally, in \secref{sec:nav}, we will demonstrate how the proposed method can
improve navigation safety.

\subsection{Simulation experiment}
\label{sec:sim}

In order to evaluate the proposed method, we  setup a simulated office
environment \cite{rasouliEffectColorSpace2017} in Gazebo. The environment, shown
in \figref{fig:office}, consists of a few different furnished zones, comprising
a meeting table, a couple of lounge areas, and an a few desks. In total, the
environment contains 19 chairs distributed among the different zones. 8 chairs
around a long meeting table, 6 chairs facing the wall with some objects in
between them, 4 office chairs facing one of the desks, and partially slid
inside, and one chair near a long table facing the room. We spawn in the
environment a simulated model of a Husky robot from Clearpath Robotics, equipped
with a Kinect RGBD camera and a SICK laser scanner.

\begin{table}
    \centering
    \caption{Experimental results}\label{tab:results}
    \begin{tabular}{lccc}
        \toprule
        Experiment       & Precision & Recall & \fscore{}\\
        \midrule
        Simulated office & 0.69      & 0.59   & 0.63\\
        Real office      & 0.71      & 0.83   & 0.78\\ 
        Real corridor    & 0.80      & 1.00   & 0.89\\
        \bottomrule
    \end{tabular}
\end{table}

The first aspect we wanted to evaluate in simulation was the ability of the
proposed method to correctly identify and then shape-complete the chairs present
in the environment. To this end, we manually teleoperated the robot to construct
a multi-layer map $\hyper{}$ of the environment using the proposed method. We
empirically set the range for $\lambda$ to $[0.1,0.25]$ m, and $\epsilon = 0.1$
m.

A total of 16 chairs were identified and shape-completed. Out of the 19 chairs
in the environment, 11 were correctly identified and the other 5 were false
positives. The first row of \tabref{tab:results} shows the achieved precision,
recall, and \fscore{}. The results are quite modest. Misdetections were caused
by inaccurate segmentations provided by Mask R-CNN, that were not fully
compensated for by the filter described in \secref{sec:completion}.
\figref{fig:false_pos} provides a top view of the scene after object replacement
and the marked objects denote the false positives, and \figref{fig:mm_tab} zooms
over the meeting table, where most of the chairs have been completed properly
except for one false positive arising from one label spreading on the table top.
Mask R-CNN represents current \sota{} capabilities in terms of object
segmentation, the results presented here are expected to improve with better
object segmentation.

\begin{figure}
	\centering
	\begin{subfigure}[b]{0.49\linewidth}
	\centering
		\includegraphics[width=\linewidth]{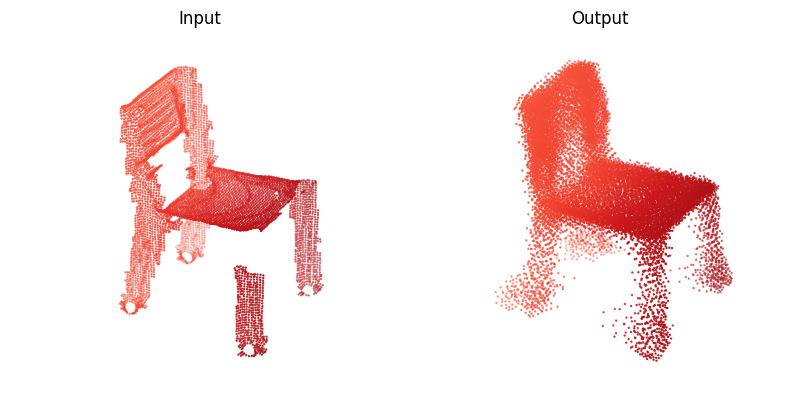}	
		\caption{Four legs}
		\label{fig:sim}
	\end{subfigure}%
	\begin{subfigure}[b]{0.49\linewidth}
	\centering
		\includegraphics[width=\linewidth]{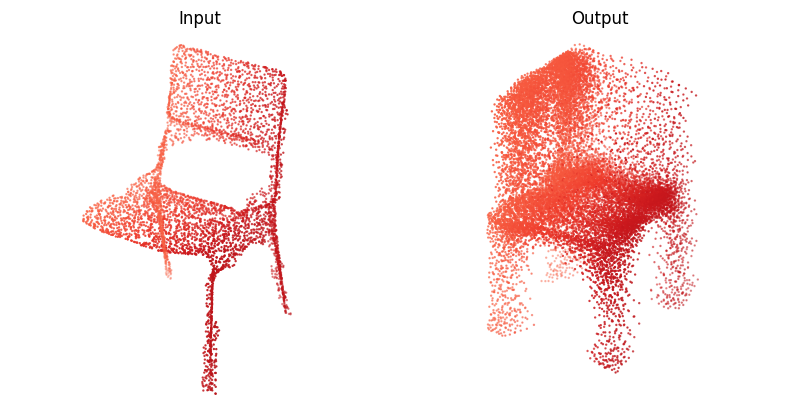}	
		\caption{Three legs}
		\label{fig:sim_3}
	\end{subfigure}
	\begin{subfigure}[b]{.49\linewidth}
	\centering
		\includegraphics[width=\linewidth]{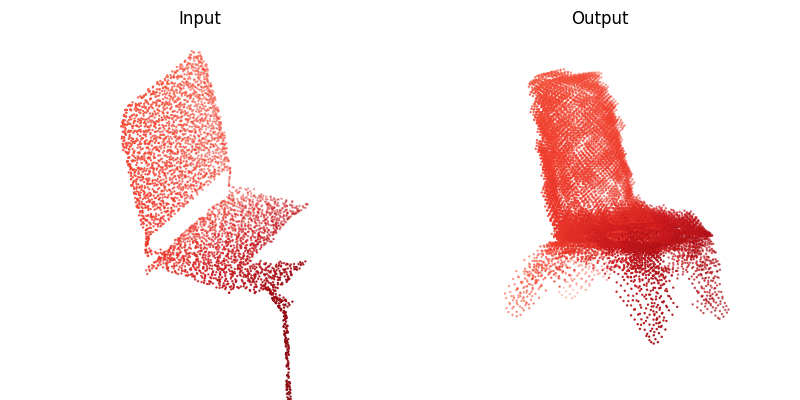}
		\caption{One leg}
		\label{fig:sim_1}		
	\end{subfigure}%
	\begin{subfigure}[b]{.49\linewidth}
	\centering
		\includegraphics[width= \linewidth]{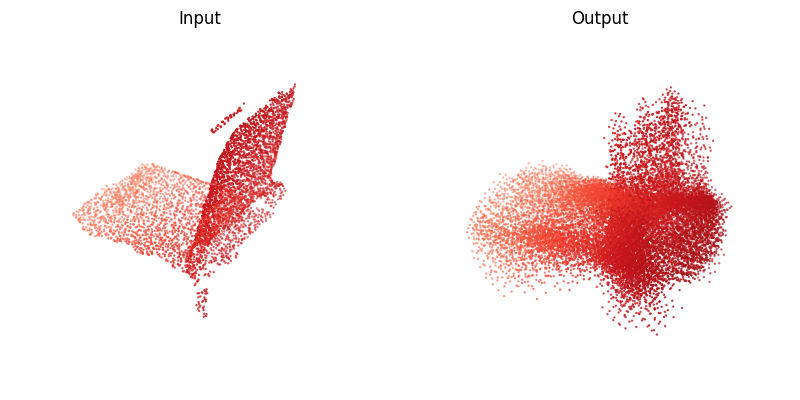}
		\caption{No legs}
		\label{fig:sim_no}
	\end{subfigure}
    \caption{Performance of \ac{pcn} on chairs with various degree of
    completion. Input (left) and output (right) are in the same pose,
    misalignment is due to a failure of the network.}\label{fig:pcn_perform}
\end{figure}

Another investigated aspect was the quality of shape-completion produced by the
proposed method, in comparison to  \sota{} deep learning methods. To this end,
we compared with \ac{pcn} \cite{yuanPCNPointCompletion2018}.
\figref{fig:pcn_perform} shows some examples of the output of \ac{pcn} for
chairs mapped at different level of completion; even when the input is lacking
only few regions, the output produced by the network has visible outliers and
the internal pose estimation of the network fails. When a similar chair in a
different pose as in \figref{fig:sim_3} is subjected to the network, the
confusion of the network is quite evident as it tries to recreate the backrest
resulting in a box-like structure above the legs. \figref{fig:sim_no} and
\ref{fig:sim_1} are the same chairs in different orientations. For the former,
the network is unable to map it to a chair, whereas for the latter it is not
able to infer the extent of legs with one partial leg. All these observations
point to the fact that the network performance does not seem to translate to
realistic partial views that can be obtained through acquisition ``in the
wild''.

\begin{figure}
	\centering
	\begin{subfigure}[b]{\linewidth}
	    \centering
		\includegraphics[width=.23\linewidth]{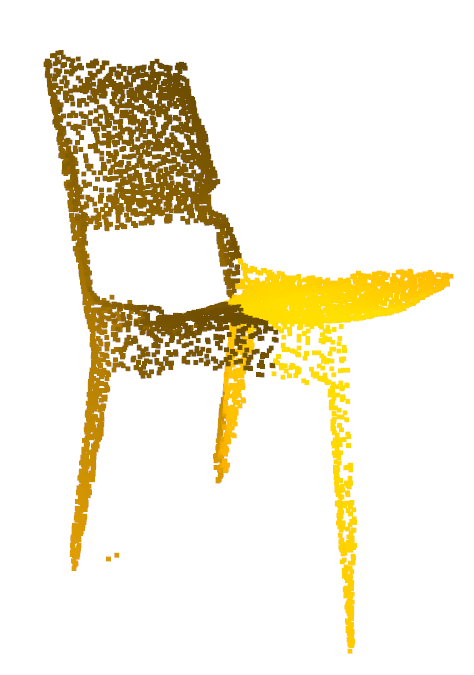}
		\includegraphics[width=.23\linewidth]{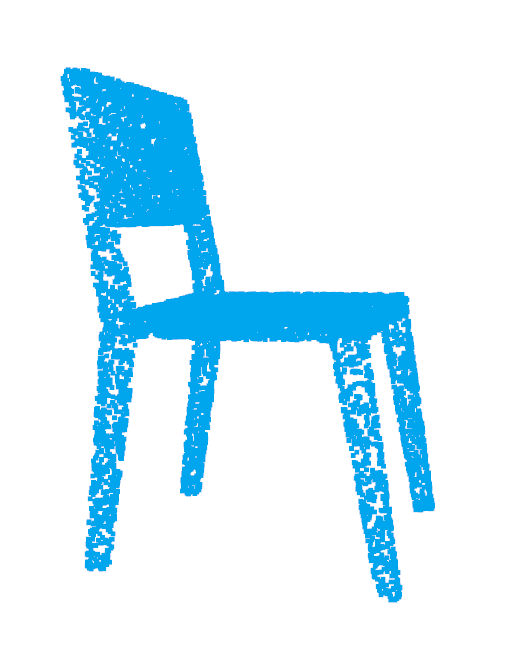}\hfill
		\includegraphics[width=.23\linewidth]{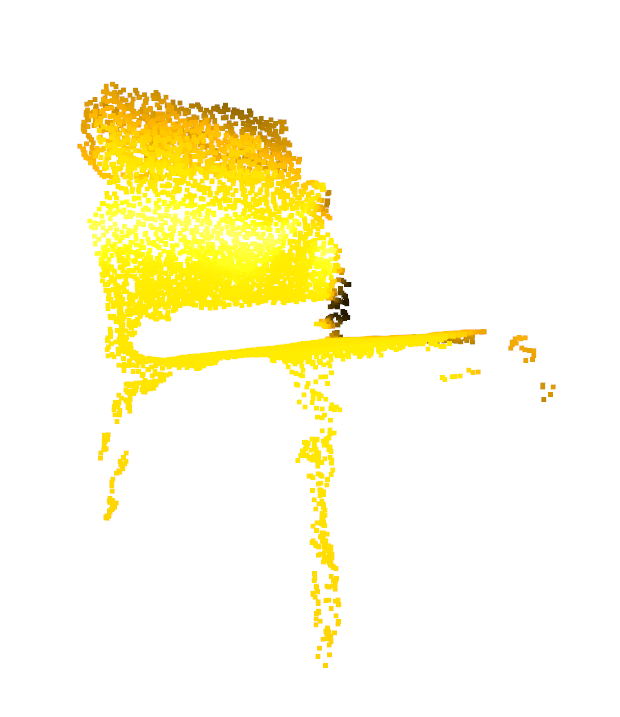}
		\includegraphics[width=.23\linewidth]{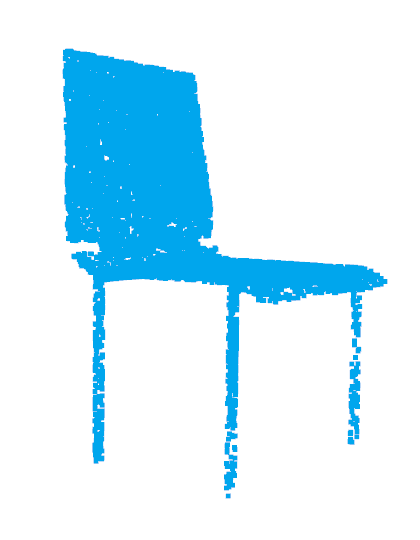}
		\caption{Input has data for backrest, seat, and
		leg/legs.}\label{fig:mm_good}
	\end{subfigure}
	
	\begin{subfigure}[b]{\linewidth}
	    \centering
		\includegraphics[width=.23\linewidth]{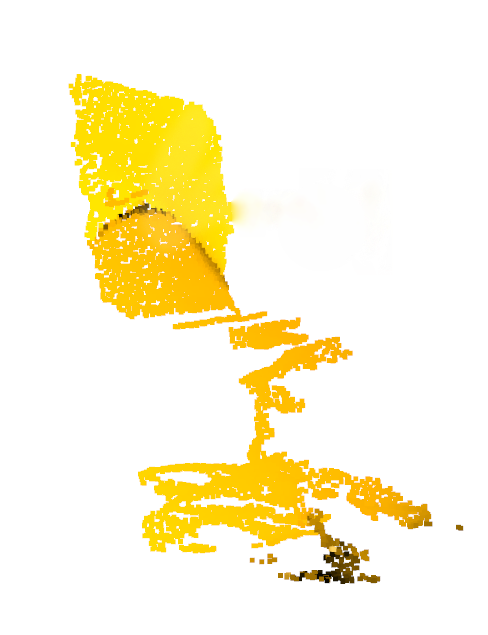}
		\includegraphics[width=.23\linewidth]{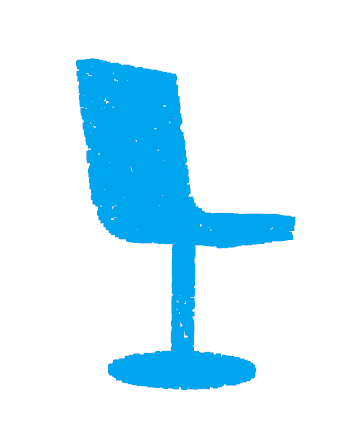}\hfill
		\includegraphics[width=.23\linewidth]{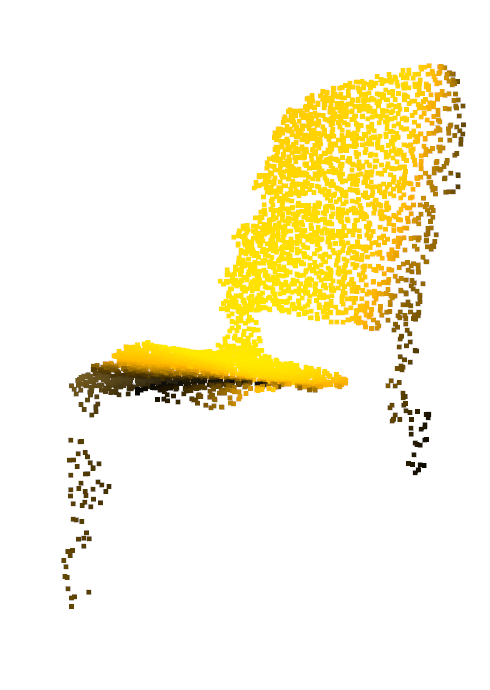}
		\includegraphics[width=.23\linewidth]{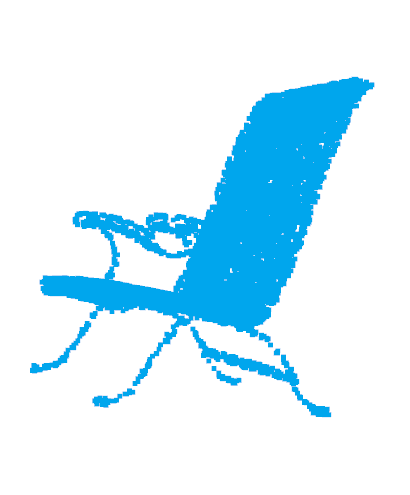}
		\caption{Input has a key component missing}\label{fig:mm_bad}
	\end{subfigure}
    \caption{Matched models using the proposed method. Input \pc{}s and model
    outputs are shown in yellow and blue respectively
    \bestcolor{}.}\label{fig:mm}
\end{figure}

The errors seen with \ac{pcn}, \ie{} noise and outliers, are not present when
using the proposed model matching system. In this case, what influences the
output quality is the region of missing information rather than its quantity
(\figref{fig:mm}). When the input has parts of all the factors that define a
chair which are legs, backrest, seat, and arms a good approximate match is
obtained as seen in \figref{fig:mm_good}. However, when a key component is
completely missing from the input, the match may differ significantly from the
actual object. In \figref{fig:mm_bad}, we can see that a wheeled office chair
lacking arms and the wheeled base in the reconstruction is completed with a
circular base instead of a wheeled base (left), while, when only a partial
extent of one leg is present (right), the corresponding match found has shorter
legs and if this model is placed back in the scene it would hover in the air.
The output also has a set of arms that are not present in the actual object.
From the observations, it is clear that a complete extent of at least one leg is
needed in the input to find a good match that would be properly grounded.
However, even when a different chair model is identified as match, the pose and
overall shape produced by the proposed method are still consistent with the
observed object, which may make these models useful for robotics applications,
as demonstrated in \secref{sec:nav}.

\subsection{Physical experiment}
\label{sec:real}

The tests on simulation provided insights into the performance, the factors
affecting it, and the bottlenecks in the pipeline. With these limitations in
mind, the viability of the pipeline was tested on a Care-O-bot 4
\cite{DBLP:conf/mc/KittmannFSRWH15} robot produced by Fraunhofer IPA. The robot
is equipped with multiple RGBD cameras on the head, neck, and torso to perceive
the area in front of it. It is capable of omnidirectional motion with a maximum
velocity of 1.1 m/s. Three 2D laser scanners mounted in the base allow the robot
to react to static and dynamic obstacles. For the experiments only the RGBD
camera mounted in the neck area was used. We mapped two environments using the
proposed method: an office where we randomly placed some chairs, and a corridor
with chairs along the wall.

\subsubsection{Office}
\label{sec:lab}

\begin{figure}
	\centering
	\begin{subfigure}[b]{.49\linewidth}
		\includegraphics[width=\linewidth]{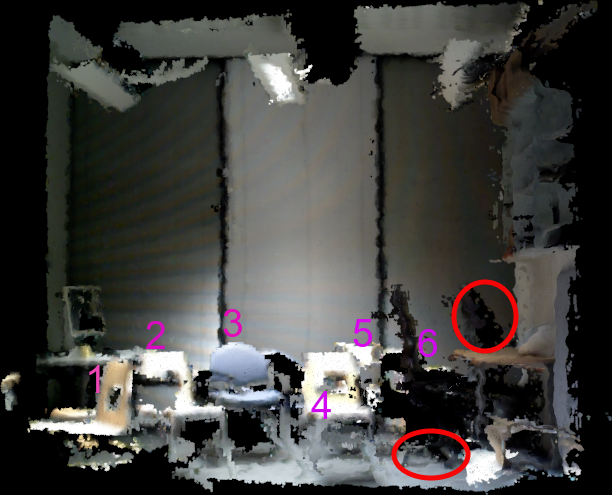}
		\caption{Office}\label{fig:real_office}
	\end{subfigure}
	\begin{subfigure}[b]{.49\linewidth}
		\includegraphics[width=\linewidth]{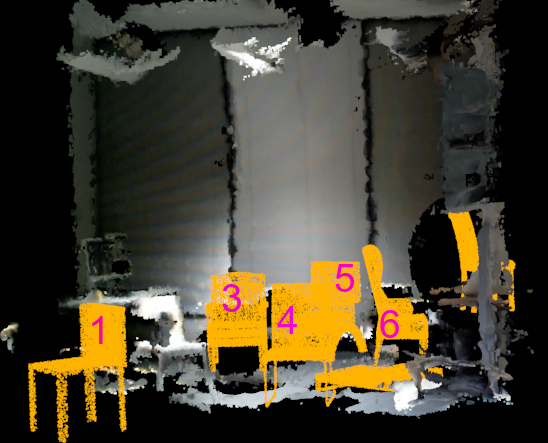}
		\caption{Office with replaced chairs}\label{fig:real_office_res}
	\end{subfigure}
	\begin{subfigure}[b]{.49\linewidth}
		\includegraphics[width=\linewidth]{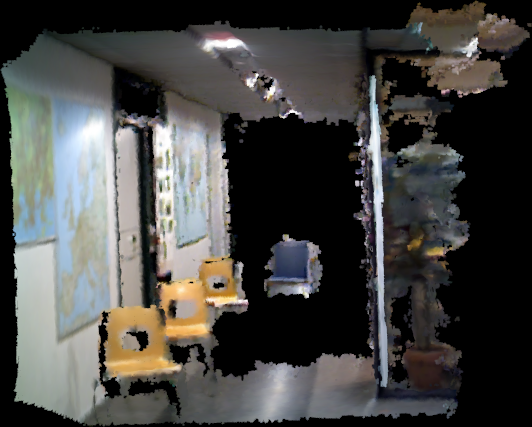}
		\caption{Corridor}\label{fig:real_corridor}
	\end{subfigure}
	\begin{subfigure}[b]{.49\linewidth}
		\includegraphics[width=\linewidth]{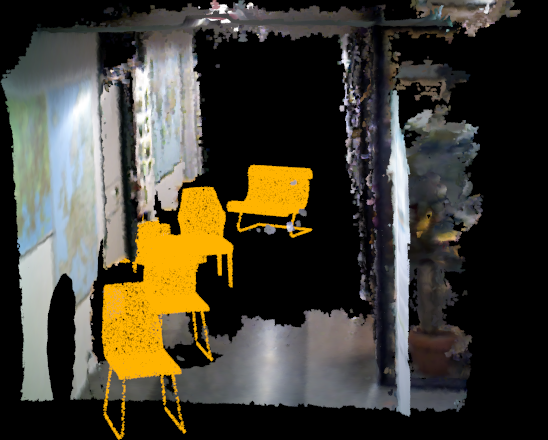}
		\caption{Corridor with replaced chairs}\label{fig:corridor_res}
	\end{subfigure}
    \caption{Pipeline performance}\label{fig:real}
\end{figure}

7 chairs of multiple designs were laid out in a chaotic manner in the room. The
robot was teleoperated through the environment for one minute and the
environment was mapped with the proposed pipeline using data captured through
the RGBD camera mounted in the neck. \figref{fig:real_office} shows the
environment as reconstructed by the pipeline (layer $\voxb{}$), while
\figref{fig:real_office_res} shows the augmented scene $\augm{}$ with the
complete object models.

As shown in \tabref{tab:results}, the setup yielded an \fscore{} of 0.78,
significantly better than the simulation. Of the five correctly detected chairs,
two (chairs 1 and 4) have good reconstruction quality, and three are reasonable
but incomplete: chair 3 because of noise in depth measurements, chair 6 because
of inaccurate semantic segmentation, and chair 5 because of heavy occlusion in
data. Two false positives were caused by inaccurate semantic segmentation of
chair 6. The wheeled base, backrest, and seat were registered whereas the stem
connecting the seat with the base was not registered, causing a false positive
seen below chair 6. Another false positive was the result of label spreading of
Mask R-CNN by which part of the wall behind chair 6 got mislabeled.

\subsubsection{Corridor}
\label{sec:corridor}

4 chairs were laid out in a narrow corridor and the robot was immobile in a
single observation position. \figref{fig:corridor_res} depicts the result of the
pipeline. The \fscore{} of the reconstruction quality was 0.89
(\tabref{tab:results}), which shows a level of performance above the previous
cases. Out of four correct detections, three were accurately reconstructed while
one was incomplete due to sensor noise. There was one false positive due to
label spreading.

In summary, the inherent shortcomings of Mask R-CNN and label spreading were
present also in the real environment. Additionally, it was observed that major
occlusions and noise in the depth information influence the quality. However,
even if the geometric shape of distant or heavily occluded chairs was
inaccurate, their pose was matched quite accurately. Correct pose estimation is
a crucial property in robotics, as we will show in the demonstration that
follows.

\subsection{Application: Shape-aware autonomous navigation}
\label{sec:nav}

\begin{figure}
	\centering
	\begin{subfigure}[b]{0.49\linewidth}
	    \centering
		\includegraphics[width=.9\linewidth]{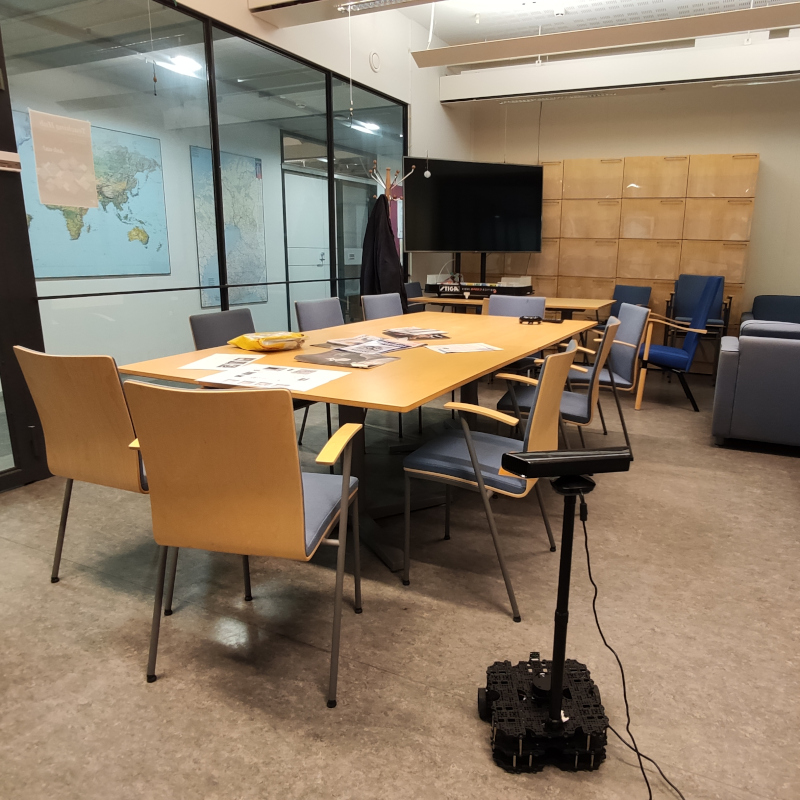}	
		\caption{Coffee room}
		\label{fig:coffeeroom}
	\end{subfigure}
	\begin{subfigure}[b]{.49\linewidth}
	    \centering
		\includegraphics[width=.9\linewidth]{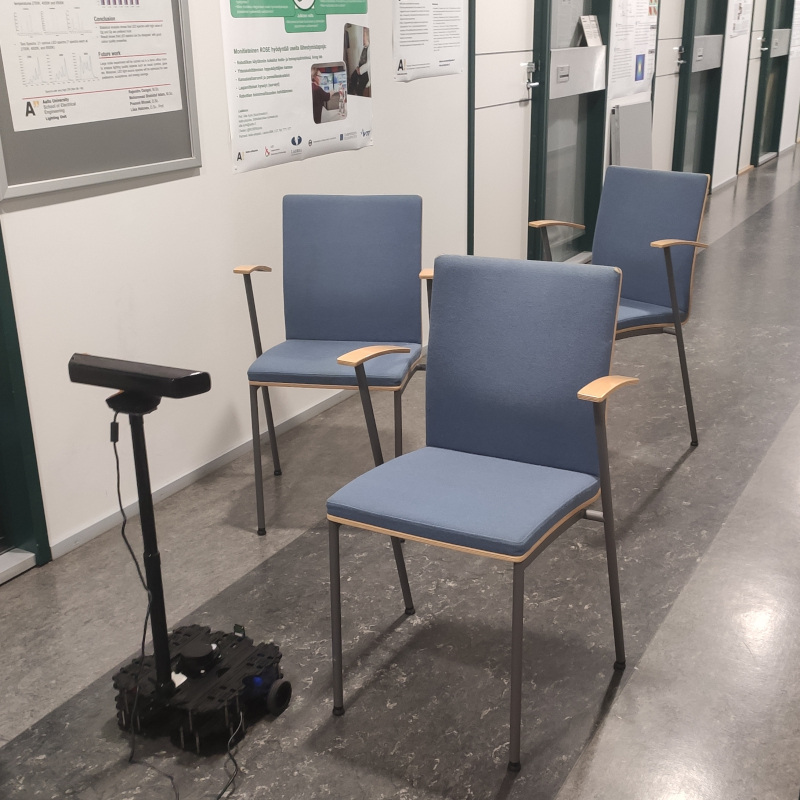}
		\caption{Corridor}
		\label{fig:corridor}		
	\end{subfigure}
    \begin{subfigure}[b]{.49\linewidth}
	    \centering
		\includegraphics[width=.9\linewidth]{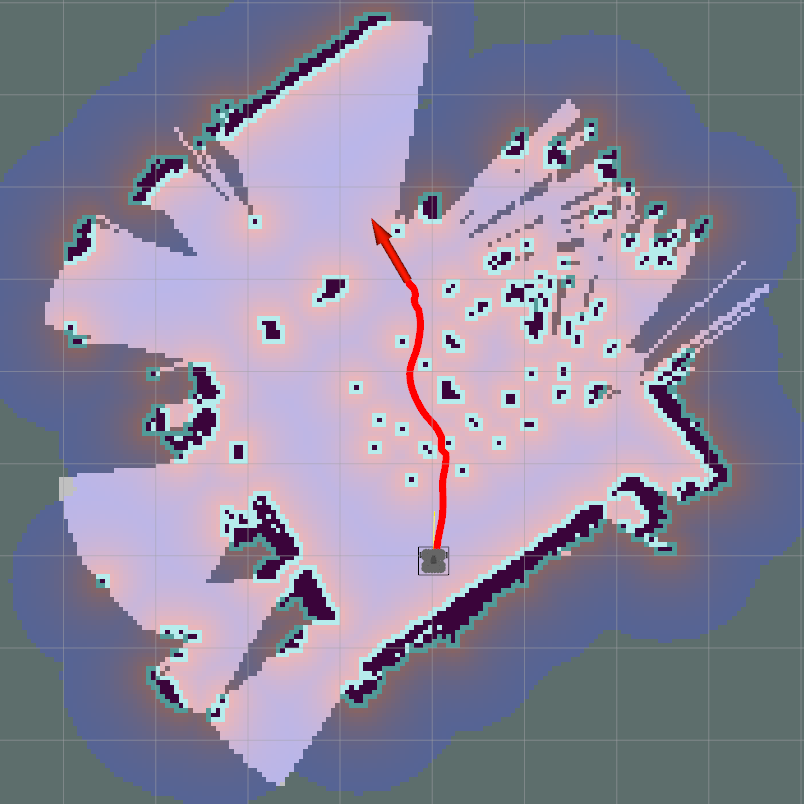}
		\caption{Coffee Room: \ac{slam}}
		\label{fig:coffeeroom_slam}		
	\end{subfigure}
	\begin{subfigure}[b]{.49\linewidth}
        \centering
    	\includegraphics[width=.9\linewidth]{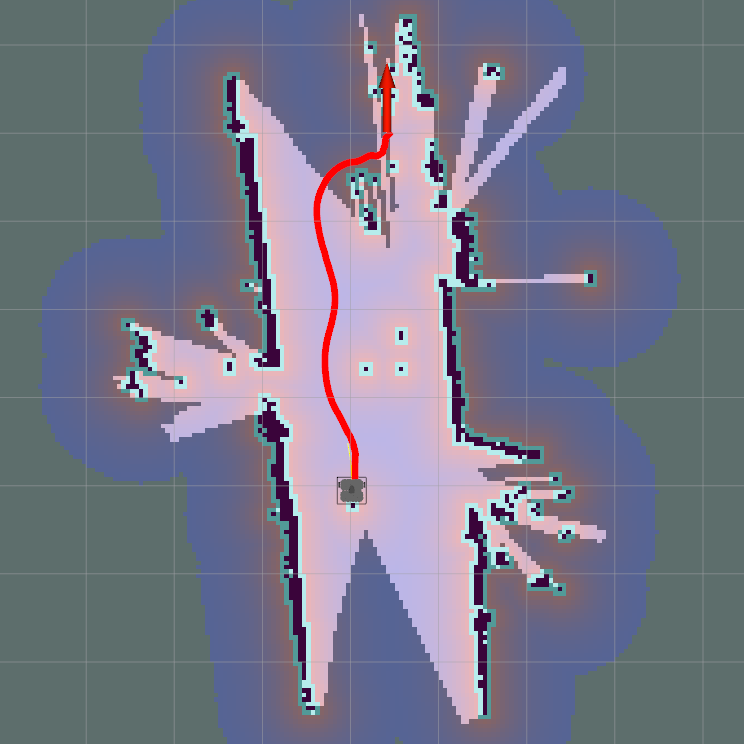}
    	\caption{Corridor: \ac{slam}}
    	\label{fig:corridor_slam}		
    \end{subfigure}
	\begin{subfigure}[b]{0.49\linewidth}
	    \centering
		\includegraphics[width=.9\linewidth]{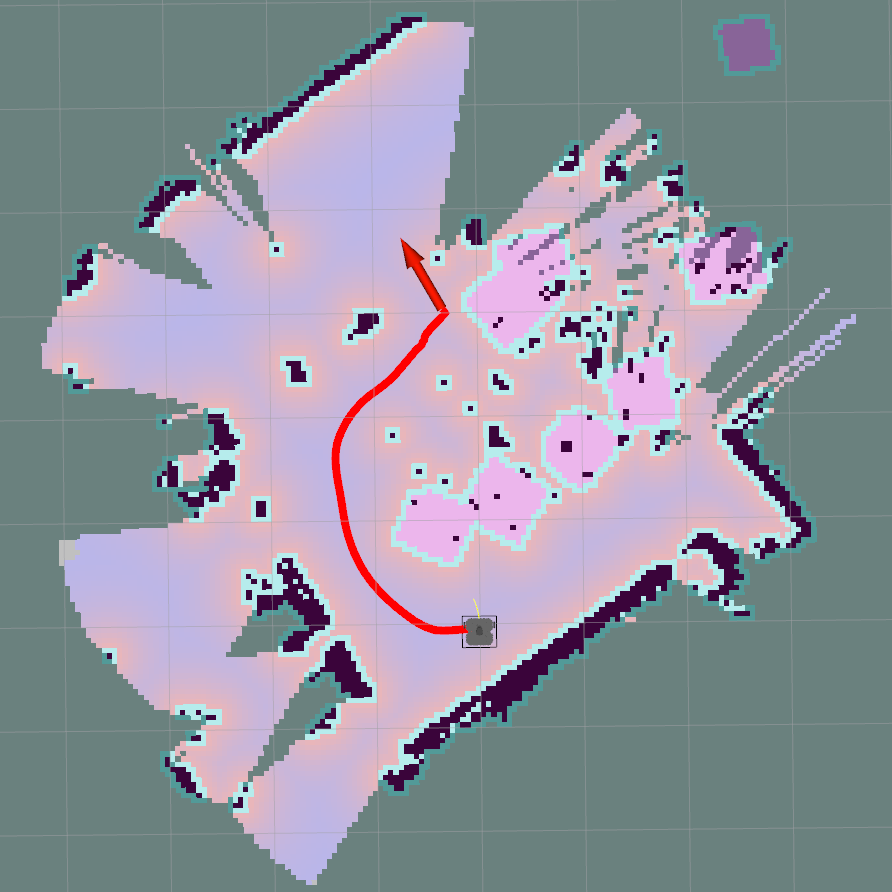}	
		\caption{Coffee Room: Multimap}
		\label{fig:coffeeroom_multimap}
	\end{subfigure}
    \begin{subfigure}[b]{.49\linewidth}
        \centering
    	\includegraphics[width=.9\linewidth]{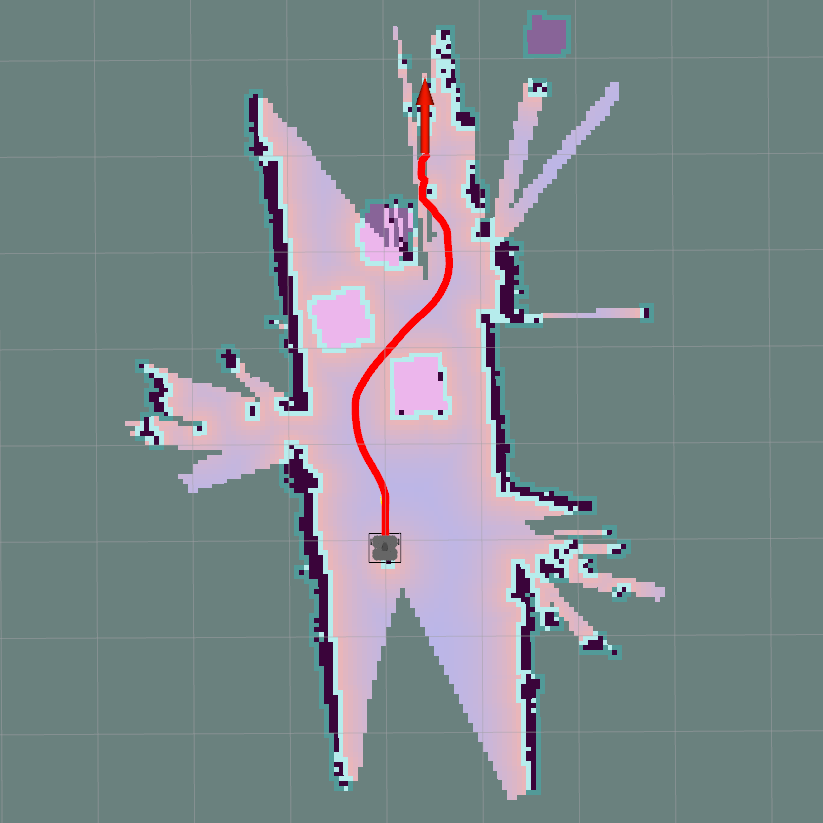}
    	\caption{Corridor: Multimap}
    	\label{fig:corridor_multimap}		
    \end{subfigure}
    \caption{Navigation experiment results over two scenarios. By navigating
    using \ac{slam} map only, \subfigref{fig:coffeeroom_slam} and
    \subfigref{fig:corridor_slam}, the robot collides with chairs, while it
    avoids all chairs by navigating on the multimap,
    \subfigref{fig:coffeeroom_multimap} and \subfigref{fig:corridor_multimap}
    \bestcolor{}.}\label{fig:navigation}
\end{figure}

As a demonstration of practical usefulness of the multi-layer mapping framework
proposed in this work, we explore its use for robotic navigation. Most robots
navigate using 2D occupancy maps built using 2D lidar-based \ac{slam}, however
those maps only represent occupancy at one specific height and fail to capture
the occupancy of complex obstacles. In this demonstration, we custom fitted a
Turtlebot3 Waffle Pi robot with a Kinect RGBD camera mounted on a short pole to
improve its field-of-view. We setup two environments, shown in
\figref{fig:coffeeroom} and \figref{fig:corridor}, where the robot is on one
side of some chairs and has to navigate to the other side. The chair seats
provide obstacles for the robot such that if the robot plans a trajectory
through the legs of a chair, the pole will collide with a seat. 

First, the robot navigated using a 2D lidar-based map and, as can be seen in
\figref{fig:coffeeroom_slam} and \figref{fig:corridor_slam}, the resulting
trajectories passed through the chairs and caused a collision. Then, estimated
object occupancies were projected down to build another 2D costmap, by
projecting all points in the object layer having $z \in [0.1, h]$, where $h$ was
the height of the robot. Using this map, the robot navigated avoiding all
obstacles and reached its destination (\figref{fig:coffeeroom_multimap} and
\figref{fig:corridor_multimap}). 

This demonstration serves first of all to illustrate the usefulness of
multi-layer maps that allow integration of information across sensors and map
layers. Secondly, it shows how the knowledge maintained by proposed mapping
framework can enable reasoning on multiple representations to increase robotic
safety, autonomy, and reliability.
\section{Discussion}
\label{sec:disc}

Over the past decade, deep learning based image segmentation and shape
completion methods have made significant leaps from a computer vision
perspective. However, our experiments hint at a problem of applying the current
methods in robotics: noisy and incomplete data captured in the wild often
deteriorates their performance considerably. This is particularly true for deep
learning methods that have usually hard time to produce reasonable results on
out-of-distribution data. In our case, the semantic label from Mask R-CNN was
often coarse which led to labels bleeding out of the object boundaries, and
required additional post processing steps. As for the methods like \ac{pcn}, the
network training is performed in isolation with existing databases without
sufficiently taking the real-world factors into account. Hence, when these
models are applied in real world robotic applications, they may fall apart as
seen in the experiments. It is important to note that while visual fidelity and
quality are important from a computer vision and graphics perspective, when
considering robotics, other factors such as precise occupancy and pose may be
more crucial.

When considering the proposed method, though it is scalable to account for more
object classes, it also adds to the requirement of extensive databases of
synthetic models for each object class and the computational overhead to search
through those databases to find a match. The process is time consuming for
larger environments, but can be performed offline. However, despite all these
factors, until deep learning methods improve their ability to transfer, our
experiments confirm model matching as the most reasonable choice for object
completion for robotics applications.
\section{Conclusions}
\label{sec:concl}

In this paper we presented a multi-layer robotic mapping pipeline able to build
in realtime geometric-semantic representation and complete object instances in
the environment by model matching. We evaluated the proposed method in both
simulation and real environments as well as compared its performance in object
shape-completion against a \sota{} deep learning method demonstrating how the
models produced by our approach are better suited for robotic applications.
Finally we demonstrated how the proposed method can be used to increase
robustness of navigation. Since the experimental scope was limited to estimating
the extent of a single object category, understanding the scalability of the
approach requires further studies. 

In conclusion, the mapping pipeline presented here represents a step toward
mapping methods, where different data sources are combined and artificial
intelligence is employed to integrate and complete missing information. More
work in this direction is still needed, particularly in regards to bridging the
gap between shape completion of synthetic object and scene completion of real
environments as well as to explore the inclusion of agent dynamics in the map.
However, we believe that such mapping approaches will be pivotal to bring the
many recent advances in computer vision to real mobile robot applications.

%%%%%%%%%%%%%%%%%%%%%%%%%%%%%%%%%%%%%%%%%%%%%%%%%%%%%%%%%%%%%%%%%%%%%%%%%%%%%%%%

\bibliographystyle{IEEEtran}
\bibliography{refs}

\end{document}